\documentclass[conference, a4paper]{IEEEtran}
\usepackage{cite}
\usepackage[pdftex]{graphicx}
\usepackage[cmex10]{amsmath}
\usepackage{algorithm}
\usepackage{algorithmicx}
\usepackage{algpseudocode}
\usepackage{mdwmath}
\usepackage{mdwtab}
\usepackage{amssymb}
\usepackage{amsmath}
\usepackage{subfigure}
\usepackage{tabularx}
\usepackage{mdwlist}
\usepackage{url}
\usepackage{multirow}

\ifCLASSINFOpdf
\else
\fi

\newcommand{\comment}[1]{}

\DeclareMathAlphabet{\mathitbf}{OML}{cmm}{b}{it}

\newcommand{\eg}{\emph{e.g.}}

\renewcommand{\Huge}{\fontsize{23.5pt}{27pt}\selectfont}
\newcommand{\sq}{\hbox{\rlap{$\sqcap$}$\sqcup$}}
\newcommand{\qed}{\hspace*{\fill}\sq}

\newcommand*{\horzbar}{\rule[.5ex]{2.5ex}{0.5pt}}

\DeclareMathOperator*{\argmin}{arg min}

\begin{document}
\title{Analyzing Zero-shot Cross-lingual Transfer\\in Supervised NLP Tasks}

\author{
\IEEEauthorblockN{Hyunjin Choi, Judong Kim, Seongho Joe, Seungjai Min, Youngjune Gwon}
\IEEEauthorblockA{Samsung SDS}
}

\maketitle

\begin{abstract}
In zero-shot cross-lingual transfer, a supervised NLP task trained on a corpus in one language is directly applicable to another language without any additional training. A source of cross-lingual transfer can be as straightforward as lexical overlap between languages (\eg, use of the same scripts, shared subwords) that naturally forces text embeddings to occupy a similar representation space. Recently introduced cross-lingual language model (XLM) pretraining brings out neural parameter sharing in Transformer-style networks as the most important factor for the transfer. In this paper, we aim to validate the hypothetically strong cross-lingual transfer properties induced by XLM pretraining. Particularly, we take XLM-RoBERTa (XLM-R) in our experiments that extend semantic textual similarity (STS), SQuAD and KorQuAD for machine reading comprehension, sentiment analysis, and alignment of sentence embeddings under various cross-lingual settings. Our results indicate that the presence of cross-lingual transfer is most pronounced in STS, sentiment analysis the next, and MRC the last. That is, the complexity of a downstream task softens the degree of cross-lingual transfer. All of our results are empirically observed and measured, and we make our code and data publicly available.
\end{abstract}

\IEEEpeerreviewmaketitle

\section{Introduction}
Pretraining language models at a large scale has dramatically improved natural language understanding. According to a comprehensive analysis~\cite{xlm-r} on the limitations in pretraining a multi-lingual model, more languages lead to better cross-lingual performance for low-resource languages only up to a certain point when the number of languages increases. The phenomenon is dubbed the curse of multilinguality, which can only be freed up by scaling up the model size. 

The recent experimental results show that multilingual models can outperform their monolingual counterparts. For a low-resource language that lacks in labeled examples, such results are an encouraging breakthrough for building an NLP application for low-resource languages. In cross-lingual language understanding, XLM by Conneau \& Lample~\cite{xlm}, despite being pretrained by only masked language modeling (MLM), has reported the state-of-the-art on downstream benchmarks. Shared lexical features (\eg, subwords, scripts, anchor points) across languages have been suspected for the primary source of learning language-independent representation that leads to cross-lingual transfer. Recent studies, however, show that parameter sharing induced by the Transformer architecture is instead the most attributable factor for the transfer.

We are motivated by these progresses in language modeling. This work focuses on empirical analysis of cross-lingual transfer in supervised NLP tasks fine-tuned over XLM. In particular, we are interested in zero-shot transfer settings where no additional training is done using the target language examples after being fine-tuned in the source language. We experiment with XLM-RoBERTa (XLM-R)~\cite{xlm-r}, a large XLM model with 550 million parameters and a 250k vocabulary size by extending semantic textual similarity, SQuAD~\cite{squad} \& KorQuAD~\cite{korquad} question answering, and sentiment classifications for various cross-lingual settings. 

At last, beyond previous work that has attempted to align word embeddings across different languages~\cite{w-mapping}, we compute a projection that directly maps sentence embeddings of one language to those of another. We then analyze the effect of fine-grained alignment of sentences across different languages to the quality of zero-shot cross-lingual transfer, manifested through the aforementioned NLP task performances measured empirically.

We make the following contributions. 
\begin{itemize}\itemsep 0pt
\item We provide rigorous results on cross-lingual transfer present in three important supervised NLP tasks that require high-level natural language understanding, namely STS, MRC, and sentiment classification. 
\item We propose to directly compute a cross-lingual mapping that aligns sentence embeddings of different languages whereas previous work has focused on word-level embeddings. 
\item We furthermore show benefits of the fine-grained cross-lingual sentence alignment that enables directly comparing sentences from different languages for sentence-pair regression tasks.
\end{itemize}

The rest of this paper is organized as follows. In Section II, we describe our approach by presenting the zero-shot cross-lingual evaluation framework. Section III discusses our experimental methodology and empirical results. Section IV concludes the paper.
\begin{figure*}[h!]
\centering
\includegraphics[width=.8\textwidth]{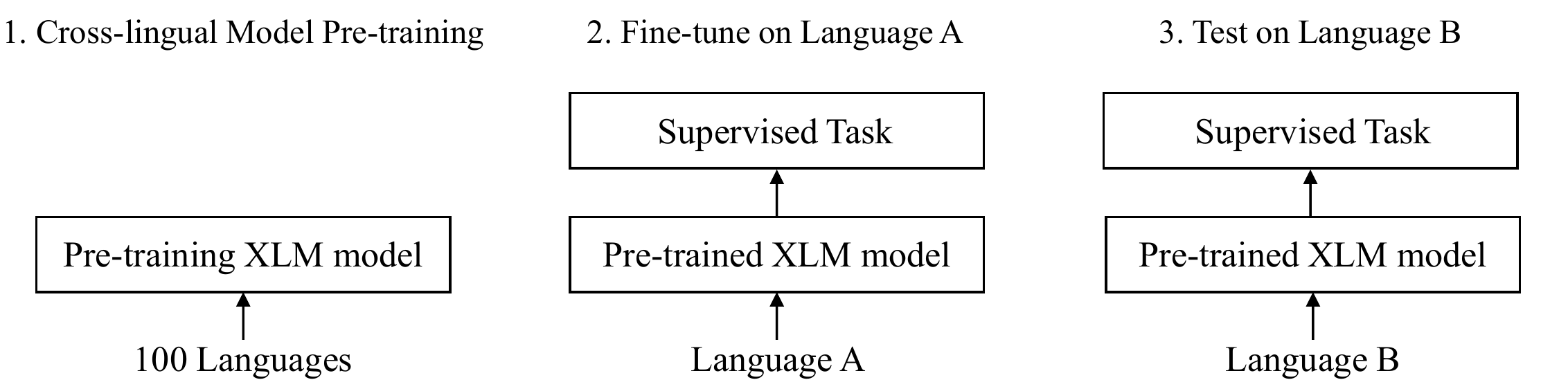}
\caption{Zero-shot cross-lingual evaluation}
\label{fig:method}
\end{figure*}

\section{Our Approach}
XLM pretraining is known to effectively promote cross-lingual transfer where a supervised model fine-tuned in one language is applied to another without additional training.

\subsection{Zero-shot Cross-lingual Evaluation Framework}
We propose a simple approach to transfer a supervised model learned in one language to another for zero-shot cross-lingual evaluation as illustrated in Fig.~\ref{fig:method}. First, we place a pretrained XLM--for our case, the 550 million-parameter XLM-RoBERTa (XLM-R)~\cite{xlm-r} trained on 100 languages is used. We then fine-tune XLM-R for a downstream task using labeled data in language A. Lastly, we evaluate the fine-tuned downstream task in both languages A and B. Note that running a test set from language B to the fine-tuned task evaluates zero-shot cross-lingual transfer. 

\subsection{Sentence Embedding and Pair Modeling}
Transformer~\cite{transformer} models such as BERT produce contextualized representations that are central to build a high-performance downstream task. XLM-R is a BERT variant whose output constitutes token embeddings (up to 512 token vectors of 768 dimensions each) for a given input. To produce fixed-size sentence embeddings necessary for a task like semantic textual similarity (STS), we average the token embedding output to obtain a single 768-dimensional pooled vector. 

\begin{figure}[h]
\centering
\includegraphics[width=.4\textwidth]{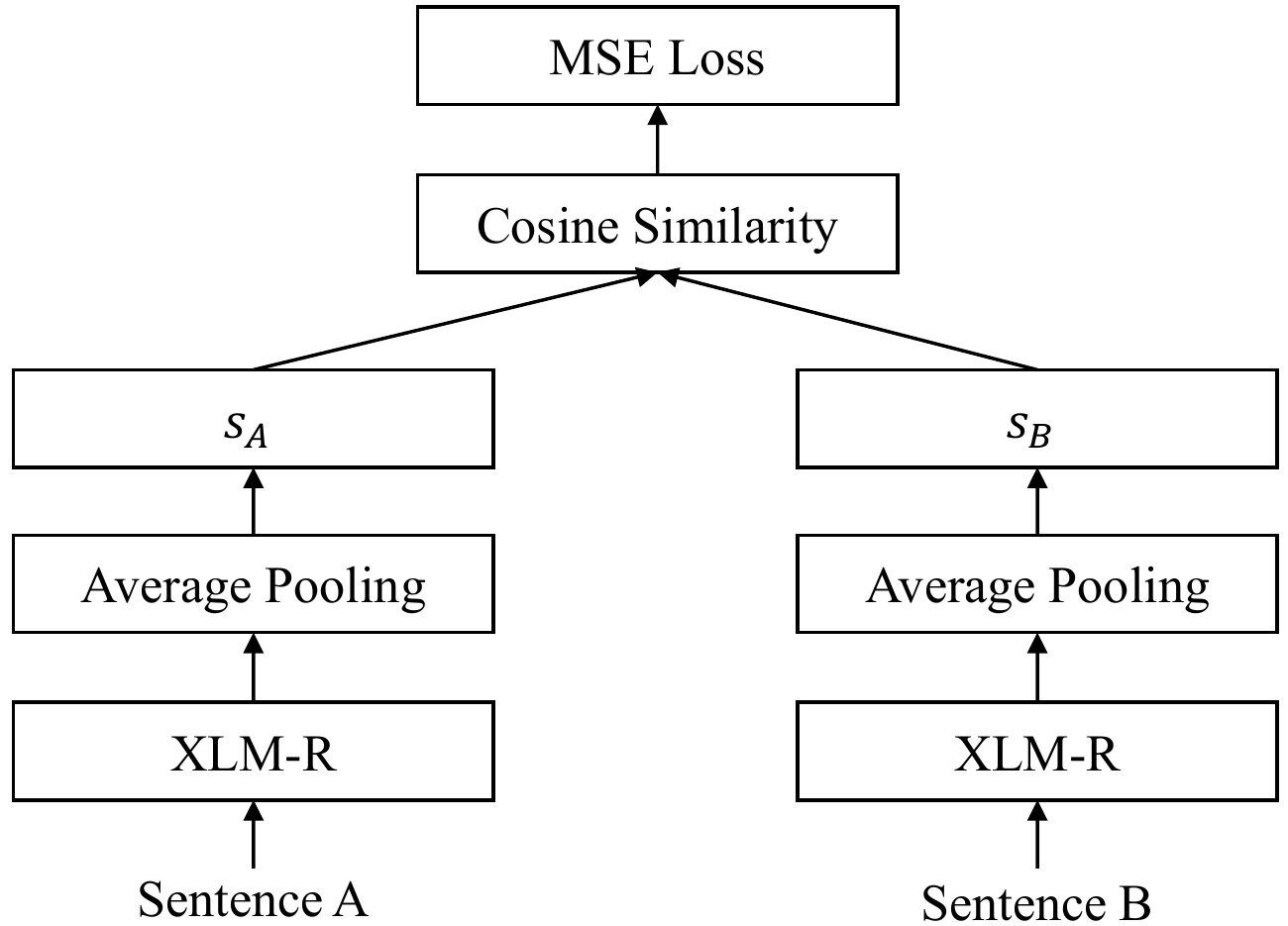}
\caption{Siamese net for sentence-pair modeling}
\label{fig:siamese}
\end{figure}

For text regression (or classification), one learns a function that maps sentence embeddings to a target value. Sentence-pair modeling gives an important primitive that underlies supervised NLP tasks such as STS. We adopt a siamese network architecture by Sentence-BERT~\cite{s-bert} that avoids the combinatorial explosion to form sentence pairs. Fig.~\ref{fig:siamese} depicts our sentence pair modeling for STS. 

\subsection{Cross-lingual Mapping for Fine-grained Alignment of Sentence Embedddings}
Cross-lingual mapping for word embeddings has been widely studied. Because context awareness is the key to language understanding, learning cross-lingual mapping for sentence-level transformations can be valuable. A sentence is less ambiguous than words since the words must be interpreted within a specific context. 

We learn cross-lingual sentence mappings directly from sentence-pair examples. Note that sentence embeddings produced from contextualized cross-lingual word embeddings would imply loosely aligned sentences. Similar to the projection-based cross-lingual word embeddings framework~\cite{w-mapping,w-mapping2}, we use linear algebraic methods to compute a projection matrix that achieves fine-grained alignment of sentence embeddings across different languages. We also use a single-layer neural net that can iteratively learn the same projection by gradient descent.

\textbf{System of least squares via normal equation.} Suppose languages $A$ and $B$ that are the source and the target languages of the projection $\mathbf{\Phi}$. We seek the solution to the problem $\mathbf{S}_A \mathbf{\Phi} = \mathbf{S}_B$ with \begin{align}\nonumber
\mathbf{S}_{A} = \begin{bmatrix}
~\horzbar ~~\mathbf{s}_{A}^{(1)} ~~\horzbar~ \\
~\horzbar ~~\mathbf{s}_{A}^{(2)} ~~\horzbar~ \\
\vdots \\
~\horzbar ~~\mathbf{s}_{A}^{(n)} ~~\horzbar~
\end{bmatrix}
,~
\mathbf{S}_{B} = \begin{bmatrix}
~\horzbar ~~\mathbf{s}_{B}^{(1)} ~~\horzbar~ \\
~\horzbar ~~\mathbf{s}_{B}^{(2)} ~~\horzbar~ \\
\vdots \\
~\horzbar ~~\mathbf{s}_{B}^{(n)} ~~\horzbar~
\end{bmatrix}
,~\\
\mathbf{s}_{A}^{(i)} =\begin{bmatrix}
a_1^{(i)}\\
a_2^{(i)}\\
\vdots \\
a_d^{(i)}
\end{bmatrix}^\top
,~ 
\mathbf{s}_{B}^{(i)} =\begin{bmatrix}
b_1^{(i)}\\
b_2^{(i)}\\
\vdots \\
b_d^{(i)}
\end{bmatrix}^\top
\end{align} where $\mathbf{S}_A$ and $\mathbf{S}_B$ are datasets that contain $n$ sentence embeddings for languages $A$ and $B$ with each sentence $\mathbf{s} \in \mathbb{R}^d$. With $\mathbf{\Phi} = \begin{bmatrix}
\boldsymbol{\phi}^{(1)} & \boldsymbol{\phi}^{(2)}  & \dots & \boldsymbol{\phi}^{(j)} & \dots & \boldsymbol{\phi}^{(d)} 
\end{bmatrix}$ whose element $\boldsymbol{\phi}^{(j)} \in \mathbb{R}^d$ is a column vector, each $\mathbf{S}_A\,\boldsymbol{\phi}^{(j)} = [b_j^{(1)} b_j^{(2)} \dots b_j^{(n)}]$ gives a problem of the least squares. Since $j = 1, \dots, d$, we have a system of $d$ least-square problems that can be solved linear algebraically via the normal equation: $\mathbf{\Phi}^* = \left(\mathbf{S}_{A}^{\top}\mathbf{S}_{A} \right)^{-1}\mathbf{S}_{A}^{\top}\mathbf{S}_{B}$.

\textbf{Solving the Procrustes problem.} Given two data matrices, a source $\mathbf{S}_A$ and a target $\mathbf{S}_B$, the orthogonal Procrustes problem~\cite{procrustes} describes a matrix approximation searching for an orthogonal projection that most closely maps $\mathbf{S}_A$ to $\mathbf{S}_B$. Formally, we write \begin{align}\label{eq:procrustes}
\mathbf{\Psi}^* = \argmin_{\mathbf{\Psi}} \left \| \mathbf{S}_A \mathbf{\Psi} - \mathbf{S}_B \right \|_{\textrm{F}}~~~~\textrm{s.t.}~\mathbf{\Psi}^\top\mathbf{\Psi} = \mathbf{I}
\end{align} The solution to Eq.~(\ref{eq:procrustes}) has the closed-form $\mathbf{\Psi}^* = \mathbf{U}\mathbf{V}^\top$ with $\mathbf{U}\mathbf{\Sigma}\mathbf{V}^\top = \textrm{SVD}(\mathbf{S}_B\mathbf{S}_A^\top)$, where SVD is the singular value decomposition. 

\begin{figure}[h]
\centering
\includegraphics[width=.4\textwidth]{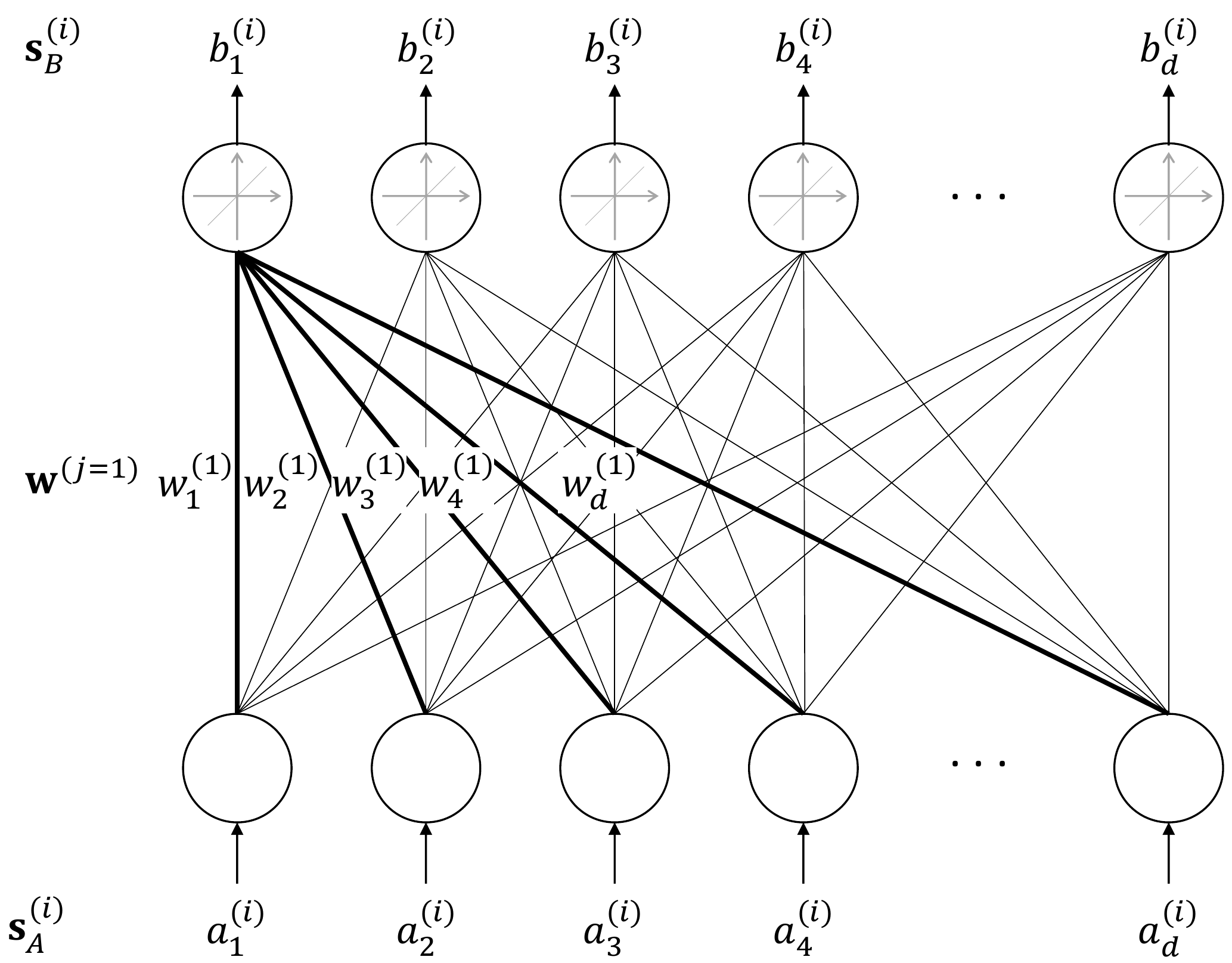}
\caption{Fully-connected single-layer neural net with neurons having linear activation functions.}
\label{fig:fcnn}
\end{figure}

\textbf{Fully-connected single-layer neural net with linearly activated neurons.} Contrasted to linear algebraic solutions $\mathbf{\Phi}^*$ and $\mathbf{\Psi}^*$, a neural net can be used to compute the projection matrix iteratively via gradient descent. We consider a fully-connected single hidden-layer neural net with linear activation functions as illustrated in Fig.~\ref{fig:fcnn}. We use the neural net as an array of linear regressors with mean square error (MSE) objectives \begin{align}
\mathbf{S}_A \mathbf{W} = \mathbf{S}_B~\textrm{(feedforward)}~\textrm{s.t.}~\frac{1}{2} \left \| \mathbf{s}_B^{(i)} - \mathbf{S}_A \mathbf{w}^{(j)} \right \|_2^2 < \epsilon~~\forall i,j
\end{align} where $\mathbf{W} = [\mathbf{w}^{(1)} \mathbf{w}^{(2)} \dots \mathbf{w}^{(j)} \dots \mathbf{w}^{(d)}]$ contains the weight parameters of the neural net. Instead of a cross-entropy loss, we impose the MSE loss function to optimize each $\mathbf{w}^{(j)}$ for stochastic gradient descent.

\begin{table*}[h!]
\begin{center}
\caption{\label{tab:sts} Evaluation on STS tasks. Numbers represent the Spearman (Pearson) correlations in percentile.}
\begin{tabular}{cl|cccc}
\multicolumn{1}{l}{\multirow{2}{*}{}}                                                                  &                           & \multicolumn{4}{c}{\textit{Evaluation Language}}                                                                                             \\
\multicolumn{1}{l}{}                                                                                   & \textit{Fine-tuning Task(s)}         & English                           & Korean                            & Spanish                           & Arabic                            \\ \hline
\multirow{2}{*}{\textit{\begin{tabular}[c]{@{}c@{}}Zero-shot\end{tabular}}} & STSb (English)            & \multicolumn{1}{l}{87.44 (87.43)} & \multicolumn{1}{l}{82.34 (82.27)} & \multicolumn{1}{l}{85.58 (87.02)} & \multicolumn{1}{l}{72.67 (70.54)} \\
                                                                                                       & KorSTS (Korean)           & \multicolumn{1}{l}{84.47 (84.40)} & \multicolumn{1}{l}{83.38 (83.16)} & \multicolumn{1}{l}{84.94 (85.00)} & \multicolumn{1}{l}{70.99 (69.66)} \\ \hline
\multirow{3}{*}{\textit{\begin{tabular}[c]{@{}c@{}}Mixed\\ Launguage\\ Fine-tuning\end{tabular}}}      & STSb $\rightarrow$ KorSTS & 86.43 (86.47)                     & 83.54 (83.42)                     & 85.47 (86.05)                     & 73.85 (73.39)                     \\
                                                                                                       & KorSTS $\rightarrow$ STSb & 88.33 (88.34)                     & 85.12 (85.12)                     & 86.77 (87.83)                     & 73.37 (72.37)                     \\
                                                                                            & STSb + KorSTS             & 87.71 (87.84)                     & 84.37 (84.48)                     & 86.53 (86.99)                     & 75.72 (75.22)                    
\end{tabular}
\end{center}
\end{table*}

\begin{table*}[h!]
\begin{center}
\caption{\label{tab:mrc} Evaluation on MRC tasks. Numbers represent F1 score, and numbers in parentheses are exact matches.}
\begin{tabular}{cl|ccc}
\multicolumn{1}{l}{}                                                                             &                              & \multicolumn{3}{c}{\textit{Evaluation Language}}                  \\
\multicolumn{1}{l}{}                                                                             & \textit{Fine-tuning Task(s)} & English       & Korean        & Spanish                            \\ \hline
\multirow{3}{*}{\textit{Zero-shot}}                                                              & SQuAD (Enlgish)              & 88.81 (81.68) & 80.92 (45.08) & 72.07 (53.18)                      \\
                                                                                                 & KorQuAD (Korean)             & 72.03 (61.93) & 89.58 (65.29) & 58.65 (43.09)                      \\
                                                                                                 & SQuAD-es (Spanish)           & 84.75 (74.51) & 78.87 (42.76) & 76.11 (59.68)                      \\ \hline
\multirow{5}{*}{\textit{\begin{tabular}[c]{@{}c@{}}Mixed\\ Language\\ Fine-tuning\end{tabular}}} & SQuAD $\rightarrow$ KorQuAD  & 85.81 (77.16) & 90.17 (66.02) & 70.54 (52.40)                      \\
                                                                                                 & SQuAD $\rightarrow$ SQuAD-es & 86.73 (76.78) & 78.16 (36.87) & \multicolumn{1}{c}{76.70 (59.87)} \\
                                                                                                 & KorQuAD $\rightarrow$ SQuAD  & 89.16 (82.20) & 88.42 (62.83) & 72.78 (53.92)                      \\
                                                                                                 & SQuAD + KorQuAD              & 84.41 (75.93) & 86.79 (62.45) & 67.72 (48.49)                      \\
                                                                                                 & SQuAD + KorQuAD + SQuAD-es   & 89.29 (81.98) & 90.41 (66.36) & 76.75 (59.66)                     
\end{tabular}
\end{center}
\end{table*}

\section{Experiments}
Throughout all our experiments, we use the pretrained XLM-RoBERTa (XLM-R)~\cite{xlm-r} downloaded from Hugging Face\footnote{https://huggingface.co/}~\cite{huggingface} unmodified, upon which we build supervised NLP tasks and fine-tune. We focus on experimenting with sentence-level representations and their cross-lingual transfer quality evaluations when used in a downstream task under zero-shot settings. There is a considerable amount of the existing literature on evaluating the cross-lingual transfer quality of word representations, which we will not cover in this paper.  

\subsection{Semantic Textual Similarity (STS)}
\textbf{Task \& dataset.} The first of our cross-lingual experiments are STS benchmark (STSb)~\cite{stsb}, Korean STS (KorSTS)~\cite{korsts}, SemEval-2017 Spanish, and SemEval-2017 Arabic. STSb is a set of English data originated for the STS task evaluations in the International Workshop on Semantic Evaluation (SemEval)~\cite{2012,2013,2014,2015,2016} between 2012 and 2017. STSb is distributed as one of the four similarity and paraphrase tasks in the GLUE benchmark~\cite{glue}. The STSb dataset includes 8,628 sentence pairs from image captions, news headlines, and user forums that are partitioned in train (5,749), dev (1,500) and test (1,379) sets. 

The STSb sentence pairs are labeled with a similarity score ranging from 0 to 5 that indicates how similar the sentences are in terms of semantic relatedness. KorSTS is a translated dataset from STSb and has exactly the same structure. SemEval-2017 Spanish and Arabic are evaluation sets from SemEval-2017 Task 1~\cite{2017}, which has 250 test pairs per each language.

\textbf{Fine-tuning.} We run the GLUE benchmark code as-is from Hugging Face to fine-tune STS tasks. This means that a text input to XLM-R is in the Sentence A--\texttt{[SEP]}--Sentence B format, which is the same as in pretraining. We use Rectified Adam (RAdam) optimizer with a linear learning rate warm-up for 10\% of the training data and a learning rate of  $4\times 10^{-5}$. We have run 4 training epochs using a batch size of 32.

To evaluate zero-shot cross-lingual transfer, we fine-tune on the STSb train set and test using the STSb, KorSTS, SemEval-2017 Spanish, and SemEval-2017 Arabic test sets, and similarly for fine-tuning and testing on KorSTS. Furthermore, we carry out the following mixed instances: 1) fine-tune on STSb the first and KorSTS the next; 2) fine-tune on KorSTS the first and STSb the next; 3) fine-tune on sentence pair examples uniformly drawn from STSb and KorSTS. 

\textbf{Results.} The upper portion of Table~\ref{tab:sts} reports the STS performances on zero-shot cross-lingual testing with 4 languages. We immediately find the presence of cross-lingual transfer strong for STS. When fine-tuned on English (the STSb train set), zero-shot testing with Korean results in 1.24\% decrease in Spearman's rank correlation. On the other hand, when fine-tuned using the KorSTS train set, zero-shot testing with English results in 3.40\% degradation.

For Spanish and Arabic, we observe better performance when fine-tuned on English. We find particularly low scores for Arabic and suspect that it is relatively lower resource language compared to the others. In fact, XLM-R uses 28.0GB of Arabic resources while for Korean 54.2GB is used, 53.3GB Spanish, and 300.8GB English~\cite{xlm-r}.

The lower portion of Table~\ref{tab:sts} shows how two-stage fine-tuning mixed with two different languages affects the performance in each language. Although the performance numbers are similar regardless of the fine-tuning order, the last language fine-tuned slightly outperforms the others.

\subsection{Machine Reading Comprehension (MRC)}
\textbf{Task \& dataset.} Reading comprehension has been one of the most challenging tasks for machine, combining natural language understanding and generation with knowledge about the world. We use Stanford Question Answering Dataset (SQuAD)~\cite{squad}, Korean Question Answering Dataset (KorQuAD)~\cite{korquad}, and Spanish SQuAD (SQuAD-es)~\cite{spanish-es} for the cross-lingual transfer evaluation of machine reading comprehension (MRC) tasks. Both SQuAD and KorQuAD consist of crowdsourced question-answer pairs from English and Korean Wikipedia articles, respectively. SQuAD-es is a translated dataset of SQuAD for Spanish.

Using SQuAD v1.1, KorQuAD v1.0, and SQuAD-es v1.1, we do the following eight cross-lingual MRC tasks. We prepare three copies of XLM-R and fine-tune them on 1) SQuAD, 2) KorQuAD and 3) SQuAD-es for testing with SQuAD (English), KorQuAD (Korean) and SQuAD-es (Spanish) dev sets. We then fine-tune cross-lingually again using 4) KorQuAD on the SQuAD fine-tuned XLM-R, 5) SQuAD-es on the SQuAD fine-tuned XLM-R, and 6) SQuAD on the KorQuAD fine-tuned XLM-R for another round of testing with the dev sets. Additionally, we fine-tune XLM-R with 7) mixed set of SQuAD and KorQuAD and 8) mixed set of SQuAD, KorQuAD, and SQuAD-es.

\textbf{Fine-tuning.} We use RAdam optimizer with a linear learning rate warm-up for 10\% of the training data and a learning rate of $2\times 10^{-5}$. We have found that running just 3 training epochs with a batch size of 48 is sufficient.

\textbf{Results.} The upper portion of Table~\ref{tab:mrc} reports the cross-lingual MRC performance evaluated on the SQuAD, KorQuAD, and SQuAD-es dev sets. For fine-tuned SQuAD, zero-shot testing with Korean and Spanish degrades 9.67\% and 5.30\% in F1 score. (Here, the compared baseline is KorQuAD dev set tested on KorQuAD train set fine-tuned XLM-R.) Fine-tuned on KorQuAD, however, zero-shot testing with English and Spanish degrades 18.89\% and 22.94\%, respectively. The results with SQuAD-es shows 4.57\% and 11.96\% decreases for English and Korean. Compared to the performance on STS tasks, the degraded performance gap measured in F1 scores and exact match is much higher for MRC tasks.

The lower portion of Table~\ref{tab:mrc} reports the cross-lingual MRC performance for mixed language fine-tuning cases. The result shows a similar trend as in STS tasks. In general, fine-tuning with an additional language seems to improve the MRC performance regardless of testing language. Fine-tuning with all other languages yields the best MRC performance as shown in the last row of Table~\ref{tab:mrc}.

\subsection{Sentiment Analysis}
\textbf{Task \& dataset.} For sentiment analysis, we use two datasets of the similar origin, namely Large Movie Review Dataset (LMRD)~\cite{lmrd} and Naver Sentiment Movie Corpus (NSMC)~\cite{nsmc}. LMRD is a movie review dataset in English. The dataset provides a set of 50,000 reviews with labels indicating whether a review is positive or negative. NSMC uses the same labeling system for movie reviews written in Korean language. The dataset consists of 200,000 reviews. Using LMRD and NSMC, we have experimented five cross-lingual evaluations: fine-tune using 1) LMRD, 2) NSMC, 3) NSMC on the LMRD fine-tuned XLM-R, 4) LMRD on the NSMC fine-tuned XLM-R, and 5) mixed set of LMRD and NSMC. All of these tasks are evaluated on the LMRD and NSMC test sets.

\textbf{Fine-tuning.} Again, using RAdam optimizer with a linear learning rate warm-up for 5\% of the training data and a learning rate of $4\times 10^{-5}$, we run 5 training epochs with a batch size of 48.

\textbf{Results.} The upper portion of Table~\ref{tab:sent} presents the zero-shot cross-lingual transfer results on sentiment analysis tasks. The numbers represent classification accuracy in percentage. Zero-shot testing with NSMC (Korean) on the LMRD fine-tuned XLM-R results in 12.05\% accuracy degradation, whereas zero-shot testing with English shows 7.63\% decrease in classification accuracy. 

The lower portion of Table~\ref{tab:sent} presents the cross-lingual sentiment analysis performance for mixed language fine-tuning cases. Here, the performance of the last language fine-tuned is improved while the first language fine-tuned degrades a little. When fine-tuned on the train set mixed with both languages, the sentiment analysis performance improves for both languages. 

\begin{table}[h]
\begin{center}
\caption{\label{tab:sent} Evaluation on sentiment classification tasks. The numbers represent classification accuracy in percentage.}
\begin{tabular}{cl|cc}
\multicolumn{1}{l}{}                                                                             &                              & \multicolumn{2}{c}{\textit{Evaluation Language}} \\
\multicolumn{1}{l}{}                                                                             & \textit{Fine-tuning Task(s)} & English                  & Korean                 \\ \hline
\multirow{2}{*}{\textit{Zero-shot}}                                                              & LMRD (English)               & 93.52                    & 79.24                  \\
                                                                                                 & NSMC (Korean)                & 86.38                    & 90.10                  \\ \hline
\multirow{3}{*}{\textit{\begin{tabular}[c]{@{}c@{}}Mixed\\ Language\\ Fine-tuning\end{tabular}}} & LMRD $\rightarrow$ NSMC      & 90.65                    & 90.12                  \\
                                                                                                 & NSMC $\rightarrow$ LMRD      & 93.69                    & 89.47                  \\
                                                                                                 & LMRD + NSMC                  & 93.80                    & 90.24                 
\end{tabular}
\end{center}
\end{table}

\subsection{Cross-lingual Mapping for Fine-grained Alignment of Sentence Embeddings}
Using the analytical findings of Section II.C, we have determined the cross-lingual mappings $\mathbf{\Phi}^*$ and $\mathbf{\Psi}^*$ linear algebraically. We have applied the mappings to align the translated sentence pairs of STSb and KorSTS. Precisely, we set the source $\mathbf{S}_A$ English sentences from STSb, and the target $\mathbf{S}_B$ Korean sentences from KorSTS. The quality of alignment via linear projections $\mathbf{\Phi}^*$ and $\mathbf{\Psi}^*$ is very similar. Based on the average cosine similarity of the translated sentence pairs, we find $\mathbf{\Phi}^*$ slightly better than $\mathbf{\Psi}^*$. 

\begin{figure*}[h]
\centering
\includegraphics[width=.7\textwidth]{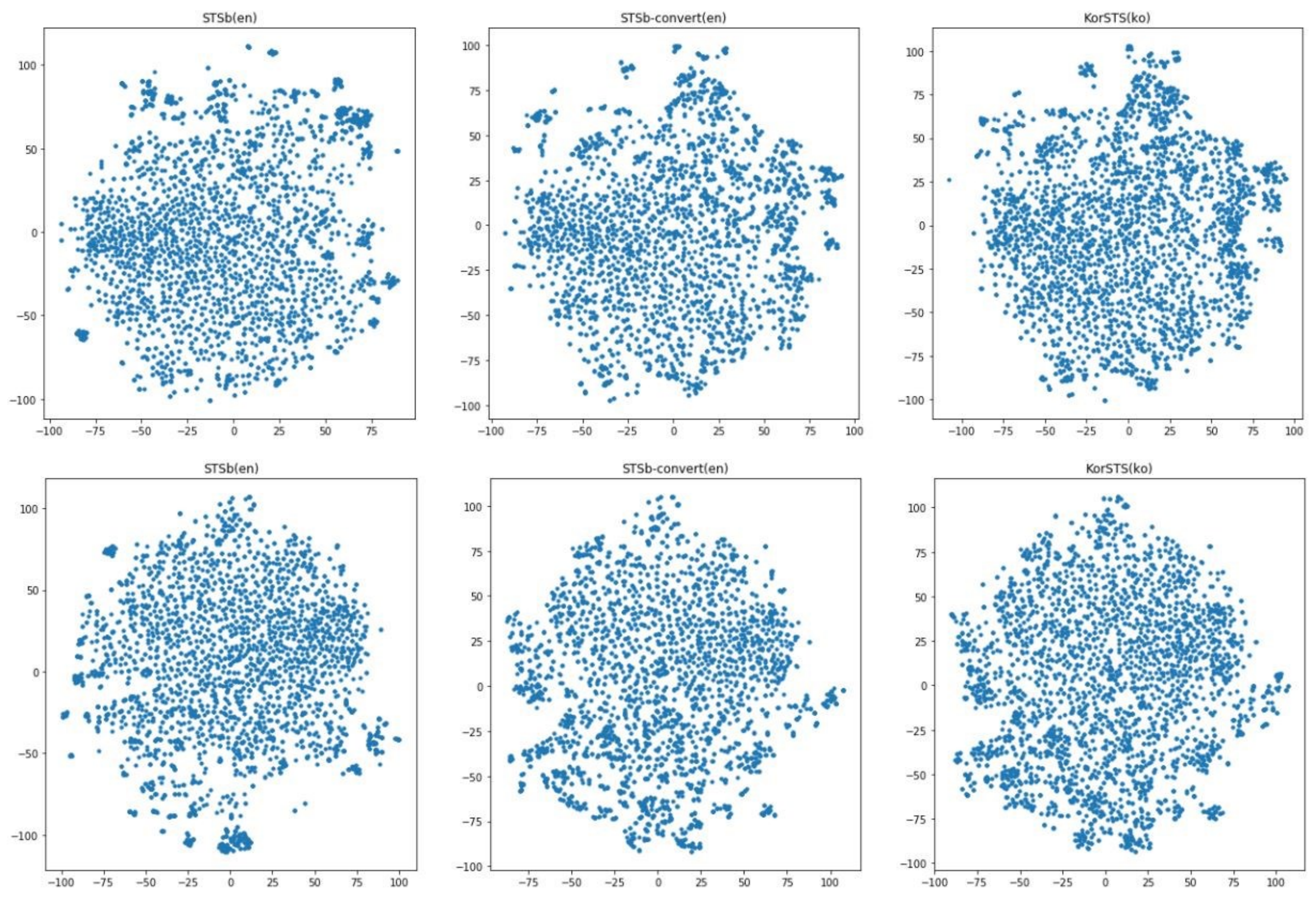}
\caption{t-SNE plots of English and Korean translated pairs from STSb and KorSTS. The leftmost plot on the top row is unaligned English sentences (source), and the middle represents aligned English via linear projection $\mathbf{\Phi}^*$, the rightmost Korean (target). The middle and the rightmost plots are aligned, showing similar patterns in t-SNE. The bottom plots are unaligned English, aligned English, and Korean sentences via the fully-connected single layer neural net whose weight parameters $\mathbf{W}$ are learned by stochastic gradient descent.}
\label{fig:plot}
\end{figure*}

We determine $\mathbf{W}$ by stochastic gradient descent on the single-layer neural net of Fig.~\ref{fig:fcnn}. Using the translated sentence pairs, we set the input $\mathbf{S}_A$ to the neural net English sentences from STSb, and the output $\mathbf{S}_B$ Korean sentences from KorSTS. The average cosine similarity of the translated sentence pairs after alignment via the $\mathbf{\Phi}^*$ projection is 0.7131 whereas the average cosine similarity for the neural net is 0.7265. Without alignment by the projection matrix or the neural net, the average cosine similarity would have been 0.4636. Fig.~\ref{fig:plot} illustrates the t-SNE plots that visualize the effect of the sentence alignment. The top plots are unaligned English, aligned English and Korean sentences by the $\mathbf{\Phi}^*$ projection, whereas the bottom plots represent unaligned and aligned English and Korean sentences via the neural net.

\begin{table}[h]
\begin{center}
\caption{\label{tab:xpair} STS evaluation with cross-lingual sentence pairs.}
\begin{tabular}{ll|cc}
                                                                                                         &        & \multicolumn{2}{c}{\textit{Method}}                                                          \\
                                                                                                         &        & \multicolumn{1}{l}{Zero-shot Transfer} & \multicolumn{1}{l}{Cross-lingual Mapping} \\ \hline
\multicolumn{1}{c}{\multirow{2}{*}{\textit{\begin{tabular}[c]{@{}c@{}}Fine-tuning\\ Task\end{tabular}}}} & STSb   & 49.03                                  & 59.16                                              \\
\multicolumn{1}{c}{}                                                                                     & KorSTS & 43.23                                  & 47.24                                             
\end{tabular}
\end{center}
\end{table}

In Table~\ref{tab:xpair}, we compare the cosine similarity of aligned English and Korean translated sentence pairs of STSb and KorSTS through the fine-grained cross-lingual mapping to zero-shot transfer. Cross-lingual mapping that we compute linear algebraically or by the use of a neural net outperforms zero-shot cross-lingual transfer by 9.3--20\% in cosine similarity matching of the translated sentences pairs of STSb and KorSTS.

\subsection{Discussion}
Generally, we find that cross-lingual transfer is present in important supervised NLP tasks that require high-level natural language understanding, namely STS, MRC, and sentiment
classification. Our empirical evaluation suggests the presence of cross-lingual transfer be most pronounced in STS. The next is sentiment analysis, and MRC comes the last. It seems that more complex a task is, and the quality of cross-lingual transfer becomes less effective. For STS, we have observed the transfer quality in two different measures, the Spearman's rank and Pearson correlation coefficients, and found them concordant. For MRC, while zero-shot transfer performance measured by F1 score is reasonable, it suffers significantly more for the case of the exact match (EM) metric. Interestingly, if we fine-tune XLM-R with both source and target languages, the last language fine-tuned has the strongest impact on the performance.

\section{Conclusion}
This paper focuses on the empirical validation of the cross-lingual transfer properties induced by XLM pretraining. We have experimented with XLM-RoBERTa (XLM-R), a large cross-lingual language model, and extended semantic textual similarity (STS), SQuAD and KorQuAD for machine reading comprehension (MRC), and sentiment analysis to cross-lingual settings. Our results suggest the presence of cross-lingual transfer be most pronounced in STS, the sentiment analysis the next, and MRC the last. We compute matrix projections linear algebraically that directly map sentence embeddings of one language to another for analyzing the effect of fine-grained alignment of sentences in zero-shot cross-lingual transfer. We have shown that such mapping can also be determined iteratively using a simple neural net. Our future work includes more systematic evaluations on broader range of low- and high-resource languages to generalize the quality of cross-lingual transfer manifested through important NLP tasks.

{
\bibliographystyle{IEEEtran}
\bibliography{paper}
}
\end{document}